\documentclass{IOS-Book-Article}

\usepackage{mathptmx}
\usepackage{soul}\setuldepth{article}

\usepackage{graphicx}
\usepackage{todonotes}
\usepackage{soul,color}
\usepackage[section]{placeins}
\usepackage{multirow}
\usepackage{bm}
\usepackage{url}
\usepackage{siunitx}
\usepackage{tikz,forest}
\usepackage{wasysym}
\usepackage{float}
\usepackage{cite}
\def\hb{\hbox to 10.7 cm{}}

\begin{document}

\pagestyle{headings}
\def\thepage{}

\begin{frontmatter}              

\newcommand{\XXX}[1]{\textcolor{red}{XXX #1}}

\title{Pretraining and Fine-Tuning Strategies for Sentiment Analysis of Latvian Tweets}

\markboth{}{July 2020\hb}

\author[A]{\fnms{Gaurish} \snm{Thakkar}%
\thanks{The work has been carried out during Gaurish's internship in Tilde.}
\thanks{Corresponding Author: Gaurish Thakkar; E-mail: gthakkar@m.ffzg.hr.}}
and
\author[B,C]{\fnms{Mārcis} \snm{Pinnis}}

\runningauthor{Gaurish and Pinnis}
\address[A]{Faculty of Humanities and Social Sciences, University of Zagreb,\\ Ul. Ivana Lučića 3, 10000, Zagreb, Croatia}
\address[B]{Tilde, Vienības gatve 75A, Riga, Latvia, LV-1004}
\address[C]{University of Latvia, Raiņa bulv. 19-125, Riga, Latvia, LV-1586}




\begin{abstract}
In this paper, we present various pre-training strategies that aid in improving the accuracy of the sentiment classification task. We, at first, pre-train language representation models using these strategies and then fine-tune them on the downstream task. 
Experimental results on a time-balanced tweet evaluation set show the improvement over the previous technique. 
We achieve {76\%} accuracy for sentiment analysis on Latvian tweets, which is a substantial improvement over previous work.

\end{abstract}

\end{frontmatter}
\markboth{July 2020\hb}{July 2020\hb}

\section{Introduction}
Sentence-level sentiment analysis (SA) aims to classify the opinion expressed by the author into either a Positive, Negative, or Neutral class. Recently, transformer-based neural networks \cite{vaswani2017attention} pre-trained using self-supervision \cite{devlin2018bert, liu2019roberta} have shown state-of-the-art performance on downstream tasks \cite{raffel2019exploring}. However, most of these findings have been reported for highly resourced languages. 

In this work, we focus on improving the performance of SA for Latvian tweets using a model of pre-trained multilingual Bidirectional Encoder Representations from Transformers (mBERT). We experiment further by pre-training the model with in-domain data. mBERT treats Unicode emoticons as out-of-vocabulary words. We propose adding them to the vocabulary of the model and repeating the pre-training and fine-tuning cycle. We also compare A-Lite-BERT (ALBERT) \cite{lan2019albert} and ELECTRA \cite{clark2020electra} models as light-weight variants of mBERT which we train from scratch on Latvian tweets. We release all the pre-trained language representation models and the models along with the code\footnote{https://github.com/thak123/bert-twitter-sentiment}. 

\section{Related Work}

Pre-trained word-embeddings \cite{mikolov2013efficient, pennington2014glove} have been studied extensively for improving sentiment classification scores \cite{severyn2015twitter}. For Latvian, Pinnis \cite{Pinnis2018} performed experiments on Latvian tweets with a wide range of classifiers and features. Peisenieks and Skadi\c{n}\v{s} \cite{peisenieks2014uses} analysed machine translation as a viable tool for performing SA for Latvian tweets. A recent study \cite{azzouza2019twitterbert} performed pre-training of the BERT model from scratch for classifying sentiment of tweet representations. Earlier techniques \cite{gulbinskis2010digitalo,vspats2016opinion} used Pointwise  Mutual Information (PMI) with Information Retrieval (IR) as well as Naive Bayes to classify multi-domain tweets.  

\section{Pre-training and Fine-tuning Strategies}
We follow 3 different pre-training strategies:
\begin{itemize}
    \item First, an existing pre-trained model trained on a large corpus is trained further on the in-domain corpus. 
    \item Second, the model trained using the previous method is trained further by adding new tokens into the existing vocabulary. Our initial experiments showed that emoticons are treated as unknowns (\textit{[UNK]}) as they are not present in the vocabulary of the pre-trained model. Since the mBERT model was trained on texts from Wikipedia, it is obvious to lack smileys in the text. We hypothesise that emoticons are sentiment-bearing tokens and hence important as features.
    \item Lastly, we pre-train (ALBERT and ELECTRA) models from scratch. The models may have vocabularies learned from the data or they may use  vocabularies from existing models. We perform this step to compare the performance of pre-trained models with the models that are trained from scratch.
\end{itemize}
Using the various annotated datasets, we perform fine-tuning on the downstream task of sentiment analysis using all the models described previously.

\section{Data}
In our experiments, we use the sentiment annotated corpora curated by Pinnis \cite{Pinnis2018}. The corpora are:
\begin{enumerate}
\item \textit{Gold} -- a corpus consisting of 6777 human-annotated Latvian tweets from the period of August 2016 till November 2016.
\item \textit{Peisenieks} -- a corpus consisting of 1178 human-annotated Latvian tweets created by Peisenieks and Skadi\c{n}\v{s} \cite{peisenieks2014uses}.
\item \textit{Auto} -- three sets of tweets from the period of August 2016 till July 2018 automatically annotated based on sentiment-identifying emoticons that are present in the tweets -- 23,685 tweets with emoticons, 23,685 tweets with removed emoticons, and 47,370 tweets with both present and removed emoticons.
\item \textit{English} -- a corpus of 45,530 various human-annotated English tweets from various sources that were machine-translated into Latvian.
\item A time-balanced evaluation set that consists of 1000 tweets from the period of August 2016 till July 2018.
\end{enumerate}

To pre-train word embeddings, we use also the Latvian tweets from the Latvian Tweet Corpus\footnote{https://github.com/pmarcis/latvian-tweet-corpus} \cite{Pinnis2018}. The corpus consists of 4,640,804 unique Latvian tweets that have been collected during the time-frame from August 2016 till March 2020.

\section{Experiments}
In this section, we describe the experimental setup for sentiment analysis. Our experimental setup consists of pre-training and fine-tuning steps. We perform the following pre-processing steps on the text:
\begin{enumerate}
\item Tokenization.
\item Removal of URLs.
\item Replacement of consecutive user mentions with a single mention.
\item User mention replacement with a placeholder (`\textit{mention$\_$i}' where the \textit{i} stands for the \textit{i}\textsuperscript{th} mention in the tweet).
\item Lower-casing of the whole tweet.
\end{enumerate}

\subsection{Pre-training}
We employ the script\footnote{https://github.com/huggingface/transformers/blob/master/examples/language{\-}modeling/run{\_}language{\_}modeling.py} available in the \textit{transformers}\footnote{https://github.com/huggingface/transformers} library to continue training the uncased version of the multilingual-BERT (mBERT) model. mBERT models 102 languages, which also include Latvian and other baltic languages. This step uses the 4.6 million unique tweets from the Latvian Tweet Corpus described above.  The corpus is split into train and eval and is pre-trained for 7 epochs.
In the case of unknown tokens, there are around 5 thousand unique \textit{[UNK]} tokens in the Gold dataset (train split only), which mainly are emoticons. Therefore, we sort the highest occurring emoticons and add 70 of them to the model vocabulary. Then, we perform one more cycle of pre-training with the new vocabulary in the network.
Using the same Latvian Tweet Corpus, we train two more models namely ALBERT and ELECTRA from scratch\footnote{https://github.com/shoarora/lmtuners/tree/master/lmtuners}. This is done by tokenizing the whole corpus and joining each of the two consecutive tweets together as examples to be trained. These are used to train a discriminator to decide if each token in the corrupted input was replaced by a generator sample or not. Both embedding{\_}size and hidden{\_}size are set to 256.

All models use the same vocabulary as that of the pre-trained uncased mBERT model, which uses sentence-piece\cite{kudo2018sentencepiece} as the method of tokenization and word splitting. We use a batch size of 16. For the pre-training step, the process was stopped once the perplexity score of $\approx$ 3 was achieved on the validation split of the dataset. 
\subsection{Fine-tuning}
For this step, We have the following pre-trained language representation models :
\begin{enumerate}
    \item mBERT - vanilla version.
    \item mBERT - pre-trained on the Latvian Tweet Corpus.
    \item mBERT - pre-trained on the Latvian Tweet Corpus plus emoticons added to the vocab.
    \item ALBERT and ELECTRA.
\end{enumerate}

For each tweet, the vector representing the [CLS] token is extracted and passed to the classification layer. All settings use a single 3-class softmax classification layer (\textit{positive, negative, neutral}) with a dropout value of 0.2. We employ a maximum sequence length of 150 and pad the shorter tweets. We randomly sample 1000 records as the validation set. Finally, we report the test accuracy on the model with the highest validation accuracy. No hyper-parameter tuning on the model is performed. Except for the emoticon augmented model, all other representation models share the same vocabulary.

\begin{figure}
\centering
\includegraphics[scale=0.5]{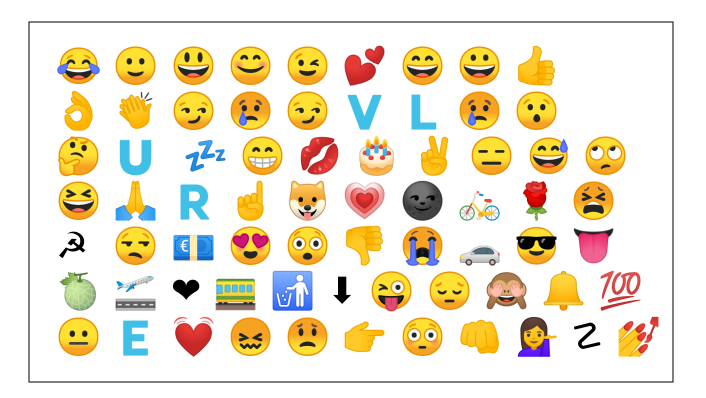}
\caption{Examples of (non-exhaustive) list of added emoticons}
\label{fig:method}
\end{figure}

\section{Results}
The accuracy results of the experiments are presented in Table~\ref{table:1}. We compare our scores with previously reported scores by \cite{Pinnis2018} (shown in the first column). The column named `Base' shows the results of using the mBERT model directly in the fine-tuning task. The next column `Pre' lists scores when the model is additionally pre-trained with in-domain Twitter data. The results show that there is an improvement over the Perceptron baseline when the mBERT model is used. There is also further improvement when we pre-train the existing model with the in-domain corpus. 
The results of the experiment where we incorporate emoticons into the vocabulary are presented in column `Pre+Emo'. This setup improves over the in-domain pre-training setup except in two cases.

The best results were obtained when the \textit{Gold} and \textit{Auto (with \smiley)} datasets were combined to train the sentiment analysis model. We see a drop in performance for all models trained on the \textit{Gold+Auto (no \smiley)} and \textit{Gold+Auto (both)} datasets.
Even though ELECTRA has a lower number of model parameters (compared to mBERT), it is still able to perform better than the vanilla mBERT version.

\begin{table}[t!]
\caption{Results of the classifier (Accuracy Scores).}
\label{table:1}
\centering
\begin{tabular}{|l|r|r|r|r|r|r|}
\hline
\multirow{2}{*}{\textbf{Dataset}}                        & \multirow{2}{*}{\textbf{Perceptron\cite{Pinnis2018}}} & \multicolumn{3}{c|}{\textbf{mBERT}}                                                 & \multirow{2}{*}{\textbf{ALBERT}} & \multirow{2}{*}{\textbf{ELECTRA}}                                 \\ \cline{3-5}
                                        & \multicolumn{1}{l|}{}                    & \multicolumn{1}{c|}{\textbf{Base}} & \multicolumn{1}{c|}{\textbf{Pre}} & \multicolumn{1}{c|}{\textbf{Pre+Emo}} &                 &                                                  \\ \hline
Gold                         &     0.661                                     & 0.678                     & \textbf{0.756}                    & 0.754                         & 0.661            & 0.711                                            \\ \hline
Gold+Peisenieks                         &        0.676                                  & 0.692                     & 0.747                    & \textbf{0.764 }                        & 0.698            & 0.706                                            \\ \hline
Gold+Auto (with \smiley) &     0.624              & 0.679                     &\textbf{ 0.769   }                 & 0.748                        & 0.649              &             0.68                \\ \hline          
Gold+Auto (no \smiley)   &       0.512                                   & 0.523                     & 0.648                    &\textbf{ 0.660 }                        & 0.483           & 0.621  \\ \hline 
Gold+Auto (both)                        &       0.487                                   & 0.526                     & 0.618                    & \textbf{0.657}                         & 0.509           & 0.564                                            \\ \hline
Gold+English                            &            0.613                              & 0.698                     & 0.692                    & \textbf{0.720   }                      & 0.669            & 0.684                                            \\ \hline
\end{tabular}
\end{table}

\section{Error Analysis}
\begin{figure}[ht!]
  \includegraphics[width=\linewidth]{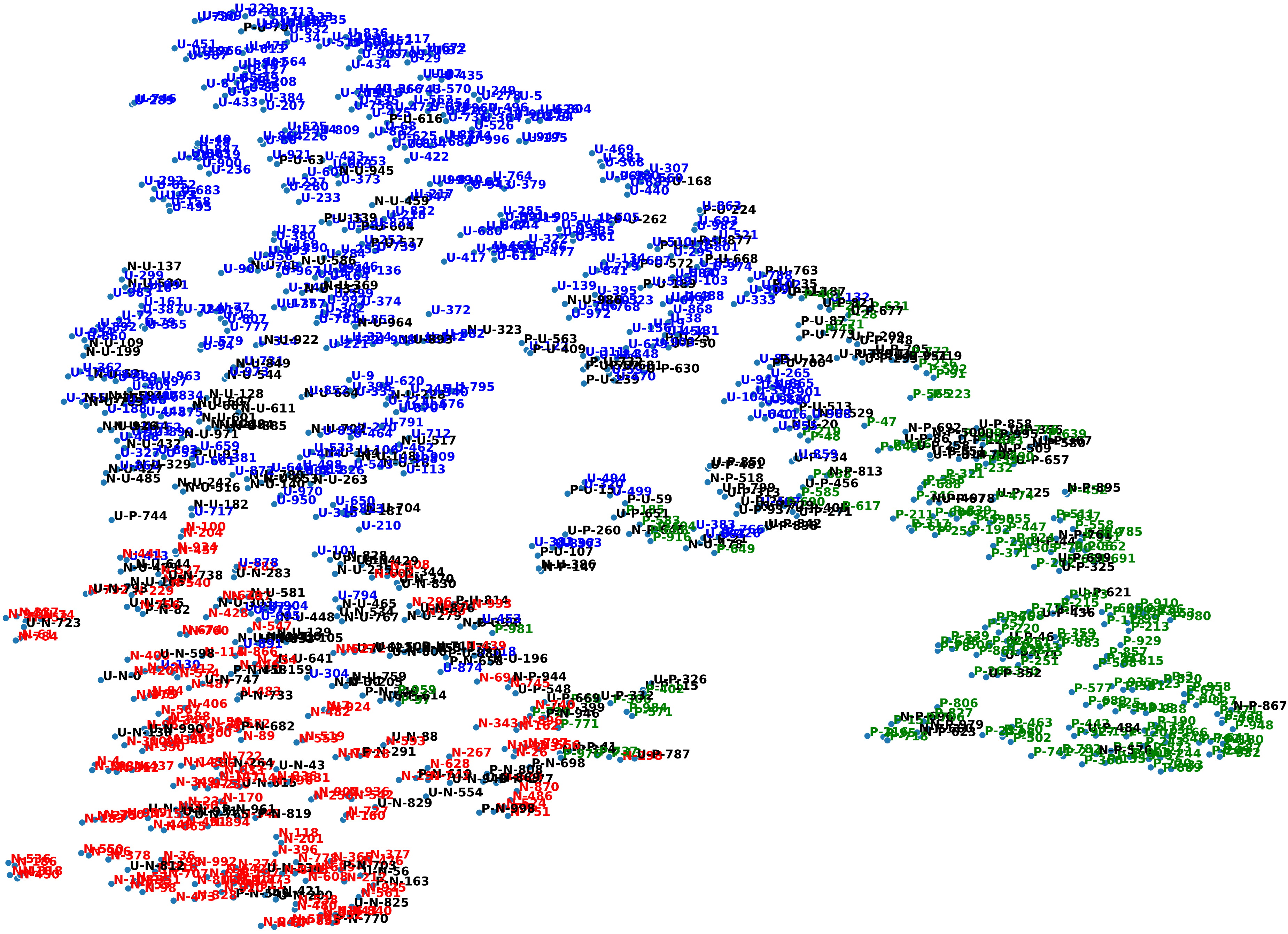}
  \caption{Tweet representation and prediction scatter plot}
  \label{fig:scatter-plot}
\end{figure}

We performed error analysis using the best-performing sentiment analysis model (i.e., the mBERT model that was additionally pre-trained on the Latvian Tweet Corpus and fine-tuned using the \textit{Gold+Auto(with \smiley)} corpus. To aid the error analysis, we visualised the test set by plotting the individual tweet representations and their predictions as scatter plots. For every tweet in the test set, we use the \textit{[CLS]} token, which is a vector of length 768, and project it down to 50 dimensions using Principal Component Analysis (PCA)\cite{wold1987principal}. The principal components are further reduced to 2 dimensions using t-SNE\cite{maaten2008visualizing}. Each of the points is plotted as nodes. We color each of the correctly predicted tweets in green for positive, red for negative, and blue for neutral. The incorrect predictions are colored in black with the correct class and predicted class as the node text. The visualisation is depicted in Figure~\ref{fig:scatter-plot}.

We started the error analysis by investigating whether we can identify if messages that are grouped in clusters that are formed in the tweet representation and prediction scatter plot have common characteristics (e.g., common syntactic structures). This did not yield positive results as the messages close to each other often contained different syntactic structures and even different vocabularies. Therefore, we continued by analysing what common characteristics can be found among a random subset of 100 misclassified tweets. From the analysis, we made the following observations:
\begin{itemize}
\item 32 of the misclassified tweets were ambiguous to the extent where external world knowledge would be necessary to decide on the polarity of the messages.
\item 17 of the misclassified tweets featured words of the wrongly selected polarity within the messages, which may indicate that the model may have learned to use lexical polarity-identifying cues to aid classification. However, it would require further analysis to validate this hypothesis.
\item 13 of the misclassified tweets featured sarcastic expressions within the messages. All of the sarcastic tweets were negative tweets. This amounts to almost 50\% of all misclassified negative tweets.
\item 12 of the misclassified tweets featured possible multiple polarities within the messages.
\item 4 tweets featured double negation.
\item 3 tweets featured spelling mistakes and a lack of diacritics that could have triggered misclassification.
\item For the remaining 19 tweets, we did not identify common characteristics.
\end{itemize}

\section{Conclusion}
In this paper, we presented our work on improving Latvian SA for tweets. Our experiments allowed us to achieve the increase in performance when pre-training word embedding models with in-domain unlabelled data and fine-tuning the models on relatively small supervised datasets. The results surpass previous work on SA for Latvian.
As future work, handling tweets with mixed emotions will be investigated. Furthermore, error analysis indicated that a large proportion of misclassified tweets can be attributed to ambiguous and sarcastic tweets for which analysis and consideration of tweet history could potentially allow expanding the context available for classification and thereby allow performing better-informed classification. Error analysis also raised a hypothesis that the fine-tuned models may have learned to focus on lexical polarity-identifying cues when deciding on which class to assign to tweets. This needs to be validated in further research. Lastly, there are still avenues of improvements to ELECTRA model pre-training evident that have not been explored and could be investigated in future work.


\section{Acknowledgments}
The  work  presented  in  this  paper  has  received  funding  from  the  European Union’s Horizon 2020 research and innovation programme under the \textit{Marie Sklodowska-Curie} grant  \textit{agreement no. 812997} and under the name CLEOPATRA (Cross-lingual Event-centric Open Analytics Research Academy). This research has been supported by the European Regional Development Fund within the joint project of SIA TILDE and University of Latvia “Multilingual Artificial Intelligence Based Human Computer Interaction” No. 1.1.1.1/18/A/148.

\bibliographystyle{ios1}           
\bibliography{ios-book-article}        

\begin{thebibliography}{17}
\ifx \bisbn   \undefined \def \bisbn  #1{ISBN #1}\fi
\ifx \binits  \undefined \def \binits#1{#1} \fi
\ifx \bauthor  \undefined \def \bauthor#1{#1} \fi
\ifx \bjtitle  \undefined \def \bjtitle#1{\textit{#1}}\fi
\ifx \batitle  \undefined \def \batitle#1{#1} \fi
\ifx \bctitle  \undefined \def \bctitle#1{#1} \fi
\ifx \bvolume  \undefined \def \bvolume#1{\textbf{#1}}\fi
\ifx \byear  \undefined \def \byear#1{#1} \fi
\ifx \bissue  \undefined \def \bissue#1{#1} \fi
\ifx \bfpage  \undefined \def \bfpage#1{#1} \fi
\ifx \blpage  \undefined \def \blpage #1{#1} \fi
\ifx \burl  \undefined \def \burl#1{#1} \fi
\ifx \doiurl  \undefined \def \doiurl#1{#1} \fi
\ifx \betal  \undefined \def \betal{et al.} \fi
\ifx \binstitute  \undefined \def \binstitute#1{#1} \fi
\ifx \beditor  \undefined \def \beditor#1{#1} \fi
\ifx \bpublisher  \undefined \def \bpublisher#1{#1} \fi
\ifx \bbtitle  \undefined \def \bbtitle#1{\textit{#1}} \fi
\ifx \bedition  \undefined \def \bedition#1{#1} \fi
\ifx \bseriesno  \undefined \def \bseriesno#1{#1} \fi
\ifx \blocation  \undefined \def \blocation#1{#1} \fi
\ifx \bsertitle  \undefined \def \bsertitle#1{#1} \fi
\ifx \bsnm \undefined \def \bsnm#1{#1} \fi
\ifx \bsuffix \undefined \def \bsuffix#1{#1} \fi
\ifx \bparticle \undefined \def \bparticle#1{#1} \fi
\ifx \barticle \undefined \def \barticle#1{#1} \fi
\ifx \botherref \undefined \def \botherref #1{#1} \fi
\ifx \url \undefined \def \url#1{#1} \fi
\ifx \bchapter \undefined \def \bchapter#1{#1} \fi
\ifx \bbook \undefined \def \bbook#1{#1} \fi
\ifx \bcomment \undefined \def \bcomment#1{#1} \fi
\ifx \oauthor \undefined \def \oauthor#1{#1} \fi
\ifx \citeauthoryear \undefined \def \citeauthoryear#1{#1} \fi
\ifx \texttildelow  \undefined \def \texttildelow{\symbol{126}} \fi
\def \endbibitem {}
\ifx \bconflocation  \undefined \def \bconflocation#1{#1} \fi

\bibitem{vaswani2017attention}
\begin{bchapter}
\bauthor{\binits{A.}~\bsnm{Vaswani}},
\bauthor{\binits{N.}~\bsnm{Shazeer}},
\bauthor{\binits{N.}~\bsnm{Parmar}},
\bauthor{\binits{J.}~\bsnm{Uszkoreit}},
\bauthor{\binits{L.}~\bsnm{Jones}},
\bauthor{\binits{A.N.}~\bsnm{Gomez}},
\bauthor{\binits{{\L}.}~\bsnm{Kaiser}} and
\bauthor{\binits{I.}~\bsnm{Polosukhin}},
\bctitle{Attention is all you need},
in: \bbtitle{Advances in neural information processing systems},
\byear{2017},
pp.~\bfpage{5998}--\blpage{6008}.
\end{bchapter}
\endbibitem

\bibitem{devlin2018bert}
\begin{botherref}
\oauthor{\binits{J.}~\bsnm{Devlin}},
\oauthor{\binits{M.-W.}~\bsnm{Chang}},
\oauthor{\binits{K.}~\bsnm{Lee}} and
\oauthor{\binits{K.}~\bsnm{Toutanova}},
Bert: Pre-training of deep bidirectional transformers for language
  understanding,
\textit{arXiv preprint arXiv:1810.04805}
(2018).
\end{botherref}
\endbibitem

\bibitem{liu2019roberta}
\begin{botherref}
\oauthor{\binits{Y.}~\bsnm{Liu}},
\oauthor{\binits{M.}~\bsnm{Ott}},
\oauthor{\binits{N.}~\bsnm{Goyal}},
\oauthor{\binits{J.}~\bsnm{Du}},
\oauthor{\binits{M.}~\bsnm{Joshi}},
\oauthor{\binits{D.}~\bsnm{Chen}},
\oauthor{\binits{O.}~\bsnm{Levy}},
\oauthor{\binits{M.}~\bsnm{Lewis}},
\oauthor{\binits{L.}~\bsnm{Zettlemoyer}} and
\oauthor{\binits{V.}~\bsnm{Stoyanov}},
Roberta: A robustly optimized bert pretraining approach,
\textit{arXiv preprint arXiv:1907.11692}
(2019).
\end{botherref}
\endbibitem

\bibitem{raffel2019exploring}
\begin{botherref}
\oauthor{\binits{C.}~\bsnm{Raffel}},
\oauthor{\binits{N.}~\bsnm{Shazeer}},
\oauthor{\binits{A.}~\bsnm{Roberts}},
\oauthor{\binits{K.}~\bsnm{Lee}},
\oauthor{\binits{S.}~\bsnm{Narang}},
\oauthor{\binits{M.}~\bsnm{Matena}},
\oauthor{\binits{Y.}~\bsnm{Zhou}},
\oauthor{\binits{W.}~\bsnm{Li}} and
\oauthor{\binits{P.J.}~\bsnm{Liu}},
Exploring the limits of transfer learning with a unified text-to-text
  transformer,
\textit{arXiv preprint arXiv:1910.10683}
(2019).
\end{botherref}
\endbibitem

\bibitem{lan2019albert}
\begin{botherref}
\oauthor{\binits{Z.}~\bsnm{Lan}},
\oauthor{\binits{M.}~\bsnm{Chen}},
\oauthor{\binits{S.}~\bsnm{Goodman}},
\oauthor{\binits{K.}~\bsnm{Gimpel}},
\oauthor{\binits{P.}~\bsnm{Sharma}} and
\oauthor{\binits{R.}~\bsnm{Soricut}},
Albert: A lite bert for self-supervised learning of language representations,
\textit{arXiv preprint arXiv:1909.11942}
(2019).
\end{botherref}
\endbibitem

\bibitem{clark2020electra}
\begin{botherref}
\oauthor{\binits{K.}~\bsnm{Clark}},
\oauthor{\binits{M.-T.}~\bsnm{Luong}},
\oauthor{\binits{Q.V.}~\bsnm{Le}} and
\oauthor{\binits{C.D.}~\bsnm{Manning}},
Electra: Pre-training text encoders as discriminators rather than generators,
\textit{arXiv preprint arXiv:2003.10555}
(2020).
\end{botherref}
\endbibitem

\bibitem{mikolov2013efficient}
\begin{botherref}
\oauthor{\binits{T.}~\bsnm{Mikolov}},
\oauthor{\binits{K.}~\bsnm{Chen}},
\oauthor{\binits{G.}~\bsnm{Corrado}} and
\oauthor{\binits{J.}~\bsnm{Dean}},
Efficient estimation of word representations in vector space,
\textit{arXiv preprint arXiv:1301.3781}
(2013).
\end{botherref}
\endbibitem

\bibitem{pennington2014glove}
\begin{bchapter}
\bauthor{\binits{J.}~\bsnm{Pennington}},
\bauthor{\binits{R.}~\bsnm{Socher}} and
\bauthor{\binits{C.D.}~\bsnm{Manning}},
\bctitle{Glove: Global vectors for word representation},
in: \bbtitle{Proceedings of the 2014 conference on empirical methods in natural
  language processing (EMNLP)},
\byear{2014},
pp.~\bfpage{1532}--\blpage{1543}.
\end{bchapter}
\endbibitem

\bibitem{severyn2015twitter}
\begin{bchapter}
\bauthor{\binits{A.}~\bsnm{Severyn}} and
\bauthor{\binits{A.}~\bsnm{Moschitti}},
\bctitle{Twitter sentiment analysis with deep convolutional neural networks},
in: \bbtitle{Proceedings of the 38th International ACM SIGIR Conference on
  Research and Development in Information Retrieval},
\byear{2015},
pp.~\bfpage{959}--\blpage{962}.
\end{bchapter}
\endbibitem

\bibitem{Pinnis2018}
\begin{bchapter}
\bauthor{\binits{M.}~\bsnm{Pinnis}},
\bctitle{{Latvian Tweet Corpus and Investigation of Sentiment Analysis for
  Latvian}},
in: \bbtitle{Human Language Technologies – The Baltic Perspective -
  Proceedings of the Seventh International Conference Baltic HLT 2018},
\bpublisher{IOS Press},
\blocation{Tartu, Estonia},
\byear{2018},
pp.~\bfpage{112}--\blpage{119}.
doi:\doiurl{10.3233/978-1-61499-912-6-112}.
\end{bchapter}
\endbibitem

\bibitem{peisenieks2014uses}
\begin{bchapter}
\bauthor{\binits{J.}~\bsnm{Peisenieks}} and
\bauthor{\binits{R.}~\bsnm{Skadins}},
\bctitle{Uses of Machine Translation in the Sentiment Analysis of Tweets.},
in: \bbtitle{Baltic HLT},
\byear{2014},
pp.~\bfpage{126}--\blpage{131}.
\end{bchapter}
\endbibitem

\bibitem{azzouza2019twitterbert}
\begin{bchapter}
\bauthor{\binits{N.}~\bsnm{Azzouza}},
\bauthor{\binits{K.}~\bsnm{Akli-Astouati}} and
\bauthor{\binits{R.}~\bsnm{Ibrahim}},
\bctitle{TwitterBERT: Framework for Twitter Sentiment Analysis Based on
  Pre-trained Language Model Representations},
in: \bbtitle{International Conference of Reliable Information and Communication
  Technology},
\binstitute{Springer},
\byear{2019},
pp.~\bfpage{428}--\blpage{437}.
\end{bchapter}
\endbibitem

\bibitem{gulbinskis2010digitalo}
\begin{botherref}
\oauthor{\binits{I.}~\bsnm{Gulbinskis}},
Digit{\=a}lo tekstu sentimenta anal{\=\i}ze
(2010).
\end{botherref}
\endbibitem

\bibitem{vspats2016opinion}
\begin{botherref}
\oauthor{\binits{G.}~\bsnm{{\v{S}}pats}} and
\oauthor{\binits{I.}~\bsnm{Birzniece}},
Opinion Mining in Latvian Text Using Semantic Polarity Analysis and Machine
  Learning Approach,
\textit{Complex Systems Informatics and Modeling Quarterly}
(2016),
51--59.
\end{botherref}
\endbibitem

\bibitem{kudo2018sentencepiece}
\begin{bchapter}
\bauthor{\binits{T.}~\bsnm{Kudo}} and
\bauthor{\binits{J.}~\bsnm{Richardson}},
\bctitle{SentencePiece: A simple and language independent subword tokenizer and
  detokenizer for Neural Text Processing},
in: \bbtitle{Proceedings of the 2018 Conference on Empirical Methods in Natural
  Language Processing: System Demonstrations},
\byear{2018},
pp.~\bfpage{66}--\blpage{71}.
\end{bchapter}
\endbibitem

\bibitem{wold1987principal}
\begin{barticle}
\bauthor{\binits{S.}~\bsnm{Wold}},
\bauthor{\binits{K.}~\bsnm{Esbensen}} and
\bauthor{\binits{P.}~\bsnm{Geladi}},
\batitle{Principal component analysis},
\bjtitle{Chemometrics and intelligent laboratory systems}
\bvolume{2}(\bissue{1--3})
(\byear{1987}),
\bfpage{37}--\blpage{52}.
\end{barticle}
\endbibitem

\bibitem{maaten2008visualizing}
\begin{barticle}
\bauthor{\binits{L.v.d.}~\bsnm{Maaten}} and
\bauthor{\binits{G.}~\bsnm{Hinton}},
\batitle{Visualizing data using t-SNE},
\bjtitle{Journal of machine learning research}
\bvolume{9}(\bissue{Nov})
(\byear{2008}),
\bfpage{2579}--\blpage{2605}.
\end{barticle}
\endbibitem

\end{thebibliography}




\end{document}